\documentclass[twocolumn, switch]{article} % Method A for two-column formatting

\usepackage{preprint}

\usepackage[numbers,square]{natbib}
\usepackage{newfloat}
\DeclareFloatingEnvironment[name={Supplementary Figure}]{suppfigure}
\usepackage{sidecap}
\sidecaptionvpos{figure}{c}

\usepackage{titlesec}
\titlespacing\section{0pt}{12pt plus 3pt minus 3pt}{1pt plus 1pt minus 1pt}
\titlespacing\subsection{0pt}{10pt plus 3pt minus 3pt}{1pt plus 1pt minus 1pt}
\titlespacing\subsubsection{0pt}{8pt plus 3pt minus 3pt}{1pt plus 1pt minus 1pt}

\title{Playing Atari with Six Neurons}

\usepackage{balance}
\usepackage{booktabs}
\usepackage{url}
\usepackage{graphicx}
\usepackage[utf8]{inputenc} % allow utf-8 input
\usepackage{booktabs}       % professional-quality tables
\usepackage{array}          % custom sizes for table columns
\usepackage{amssymb}        % extended blackboard math symbols
\usepackage{amsmath}        % complete AMS math package
\usepackage{color}          % colors
\usepackage{algorithm}      % pseudocode float
\usepackage[noend]{algpseudocode}  % pseudocode macros
\usepackage{paralist}       % compactenum

\graphicspath{{img/}}%{{../jpeg/} [...]}
\DeclareGraphicsExtensions{.pdf,.jpg,.jpeg,.png}

\makeatletter
\DeclareFontEncoding{LS1}{}{}
\DeclareFontSubstitution{LS1}{stix}{m}{n}
\DeclareMathAlphabet{\mathscr}{LS1}{stixscr}{m}{n}
\makeatother

\algnewcommand{\Inputs}[1]{%
  \State \textbf{Inputs:}
  \Statex \hspace*{\algorithmicindent}\parbox[t]{.8\linewidth}{\raggedright #1}
}
\algnewcommand{\Initialize}[1]{%
  \State \textbf{Initialize:}
  \Statex \hspace*{\algorithmicindent}\parbox[t]{.8\linewidth}{\raggedright #1}
}

\newcommand{\order}{\mathcal{O}}
\newcommand{\nparams}{\mathscr{p}}
\newcommand{\norm}{\ell}
\newcommand{\orig}{X}
\newcommand{\trainset}{\mathscr{X}}
\newcommand{\recres}{\hat{\resids}}
\newcommand{\similfn}{\text{sim}}
\newcommand{\simils}{\mathscr{S}}
\newcommand{\maxidx}{\text{index of~}\max~}
\newcommand{\resids}{\mathcal{P}}
\newcommand{\resid}{\rho}
\newcommand{\dict}{\mathcal{D}}
\newcommand{\centr}{\mathscr{d}}
\newcommand{\code}{\vec{\mathscr{c}}}
\newcommand{\append}{<\!\!<}
\newcommand{\msc}{\text{msc}} % most similar centroid
\newcommand{\ncodes}{\omega}
\newcommand{\maxncodes}{\Omega}
\newcommand{\recerr}{\mathscr{R}}
\newcommand{\recerrel}{\mathscr{r}}
\newcommand{\fit}{f}
\newcommand{\ind}{\mathbf{z}}

\newcommand{\cov}{\mathit{c}}

\author{
Giuseppe Cuccu\\
eXascale Infolab\\
Department of Computer Science\\
University of Fribourg, Switzerland\\
\texttt{name.surname@unifr.ch}
\And
Julian Togelius\\
Game Innovation Lab\\
Tandon School of Engineering\\
New York University, NY, USA\\
\texttt{julian@togelius.com}
\And
Philippe Cudré-Mauroux\\
eXascale Infolab\\
Department of Computer Science\\
University of Fribourg, Switzerland\\
\texttt{name.surname@unifr.ch}
}

\begin{document}

\twocolumn[ % Method A for two-column formatting
  \begin{@twocolumnfalse} % Method A for two-column formatting

\maketitle

\begin{abstract}
Deep reinforcement learning, applied to vision-based problems like Atari games, maps pixels directly to actions; internally, the deep neural network bears the responsibility of both extracting useful information and making decisions based on it.
By separating the image processing from decision-making, one could better understand the complexity of each task, as well as potentially find smaller policy representations that are easier for humans to understand and may generalize better.
To this end, we propose a new method for learning policies and compact state representations separately but simultaneously for policy approximation in reinforcement learning.
State representations are generated by an encoder based on two novel algorithms:
Increasing Dictionary Vector Quantization makes the encoder capable of growing its dictionary size over time, to address new observations as they appear in an open-ended online-learning context;
Direct Residuals Sparse Coding encodes observations by disregarding reconstruction error minimization, and aiming instead for highest information inclusion.
The encoder autonomously selects observations online to train on, in order to maximize code sparsity.
As the dictionary size increases, the encoder produces increasingly larger inputs for the neural network: this is addressed by a variation of the Exponential Natural Evolution Strategies algorithm which adapts its probability distribution dimensionality along the run.
We test our system on a selection of Atari games using tiny neural networks of only 6 to 18 neurons (depending on the game's controls).
These are still capable of achieving results comparable---and occasionally superior---to state-of-the-art techniques which use two orders of magnitude more neurons.
\end{abstract}
% \keywords{Game Playing \and Neuroevolution \and Evolutionary algorithms \and Learning agent capabilities} % (optional)
\vspace{0.35cm}

  \end{@twocolumnfalse} % Method A for two-column formatting
] % Method A for two-column formatting

\section{Introduction}

In deep reinforcement learning, a large network learns to map complex, high dimensional input (often visual) to actions, for direct policy approximation.
When a giant network with hundreds of thousands of parameters learns a relatively simple task (such as playing Qbert) it stands to reason that only a small part of what is learned is the actual policy.
A common understanding is that the network internally learns to extract useful information (features) from the image observation in its first layers by mapping pixels to intermediate representations, allowing the last few layer(s) to map these representations to actions.
The policy is thus learned at the same time as the intermediate representations, making it almost impossible to study the policy in isolation.

Separating the representation learning from the policy learning allows in principle for higher component specialization, enabling smaller networks dedicated to policy learning to address problems typically tackled by much larger networks.
This size difference represents a net performance gain, as larger networks can be devoted to addressing problems of higher complexity.
For example, current results on Atari games are achieved using networks of hundreds of neurons and tens of thousands of connections; making the same game playable (with comparable performance) by a network $k$ times smaller paves the road to training larger networks on $k$ independent games, using currently available methods and resources.

Separating the policy network from the image parsing also allows us to better understand how network complexity contributes to accurately representing the policy.
While vision-based tasks are often addressed with very large networks, the learned policies by themselves should in principle not require such high-capacity models, as these policies in themselves often appear to not be very complex.
Yet another reason to investigate how to learn smaller policy networks by addressing the image processing with a separate component is that smaller networks may offer better generalization.
This phenomenon is well-known from supervised learning, where smaller-capacity models tend to overfit less, but has not been explored much in reinforcement learning.

The key contribution of this paper is a new method for learning policy and features \emph{simultaneously} but \emph{separately} in a complex reinforcement learning setting.
This is achieved through two novel algorithms: Increasing Dictionary Vector Quantization (IDVQ) and Direct Residuals Sparse Coding (DRSC).

IDVQ maintains a dictionary of centroids in the observation space, which can then be used for encoding.
The two main differences with standard VQ are that the centroids are (i) trained online by (ii) disregarding reconstruction error.
Online training is achieved with the algorithm autonomously selecting images for its training from among the observations it receives to be encoded, obtained by the policies as they interact with the environment.
The disregard for reconstruction error comes instead from shifting the focus of the algorithm to the arguably more crucial criterion (from the perspective of the application at hand) of ensuring that all of the information present in the observation is represented in the centroids.
This is done by means of constructing new centroids as a residual image from the encoding while ignoring \emph{reconstruction artifacts}. See Section~\ref{subsec:compressor} for further discussion.

The dictionary trained by IDVQ is then used by DRSC to produce a compact code for each observation.
This code will be used in turn by the neural network (policy) as input to select the next action.
The code is a binary string: a value of `1' indicates that the corresponding centroid contains information also present in the image, and a limited number of centroids are used to represent the totality of the information.

As the training progresses and more sophisticated policies are learned, complex interactions with the environment result in increasingly novel observations; the dictionary reflects this by \emph{growing in size}, including centroids that account for newly discovered features.
A larger dictionary corresponds to a larger code, forcing the neural network to grow in input size.
This is handled using a specialized version of Exponential Natural Evolution Strategy which adapts the dimensionality of the underlying multivariate Gaussian.

With the goal of minimizing the network size while maintaining comparable scores, experimental results show that this approach can effectively learn both components simultaneously, achieving state-of-the-art performance on several ALE games while using a neural network of \emph{only 6 to 18 neurons}, i.e.\ \textbf{two orders of magnitude smaller} than any known previous implementation.
This research paves the road for training deep networks entirely dedicated to policy approximation, addressing problems of unprecedented complexity.

%%%%%%%%%%%%%%%%%%%%%%%%%%%%%%%%%%%%%%%%%%%%%%%%%%%%%%%%%%%%%%%%%%%%%%%%%%%%%%%%%
\begin{figure*}[!t]
\label{fig:system}
\centerline{\includegraphics[width=0.97\textwidth]{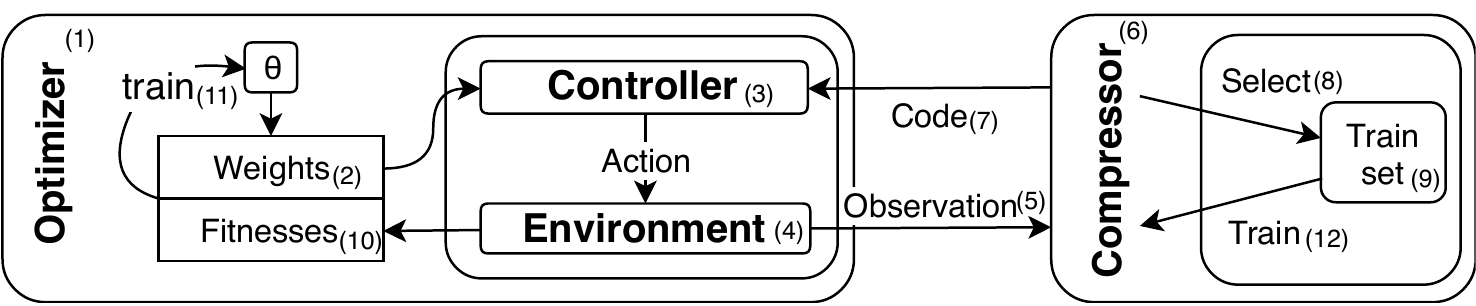}}
\caption{
  \textbf{System diagram.} At each generation the optimizer (1) generates sets of weights (2) for the neural network controller (3). Each network is evaluated episodically against the environment (4). At each step the environment sends an observation (5) to an external compressor (6), which produces a compact encoding (7). The network uses that encoding as input. Independently, the compressor selects observations (8) for its training set (9). At the end of the episode, the environment returns the fitness (cumulative reward; 10) to the optimizer for training (neuroevolution; 11). Compressor training (12) takes place in between generations.
}
\end{figure*}
%%%%%%%%%%%%%%%%%%%%%%%%%%%%%%%%%%%%%%%%%%%%%%%%%%%%%%%%%%%%%%%%%%%%%%%%%%%%%%%%%

\section{Related work}

\subsection{Video games as AI benchmarks}

Games are useful as AI benchmarks as they are designed to challenge human cognitive capacities. Board games such as Chess and Go have been used as AI benchmarks since the inception of artificial intelligence research, and have been increasingly used for testing and developing AI methods~\cite{yannakakis2018artificial}. Though various video game-based AI competitions and frameworks exist, the introduction of the Arcade Learning Environment (ALE) did much to catalyze the use of arcade games as AI benchmarks~\cite{bellemare2013arcade}.

ALE is based on an emulation of the Atari 2600, the first widely available video game console with exchangeable games, released in 1977. This was a very limited piece of hardware: 128 bytes of RAM, up to 4 kilobytes of ROM per games, no video memory, and an 8-bit processor operating at less than 2 MHz. The limitations of the original game console mean that the games are visually and thematically simple. Most ALE games feature two-dimensional movement and rules mostly triggered by sprite intersection. In the most common setup, the raw pixel output of the ALE framework is used as inputs to a neural network, and the outputs are interpreted as commands for playing the game. No fast forward model is available, so planning algorithms are ineffective. Using this setup, Mnih et al.\ reached above human level results on a majority of 57 Atari games that come with the standard distribution of ALE~\cite{mnih2015human}. Since then, a number of improvements have been suggested that have improved game-playing strength on most of these games~\cite{hessel2017rainbow,justesen2019deep}.

\subsection{Neuroevolution}

Neuroevolution refers to the use of evolutionary algorithms to train neural networks~\cite{floreano2008neuroevolution,yao1999evolving,igel2003neuroevolution,risi2017neuroevolution}. Typically, this means training the connection weights of a fixed-topology neural network, though some algorithms are also capable of evolving the topology at the same time as the weights~\cite{stanley2002evolving}.

When using neuroevolution for reinforcement learning, a key difference is that the network is only trained in between episodes, rather than at every frame or time step. In other words, learning happens between episodes rather than during episodes; this has been called \emph{phylogenetic} rather than \emph{ontogenetic} reinforcement learning~\cite{togelius2009ontogenetic}.
While it could be argued that evolutionary reinforcement learning should learn more slowly than ontogenetic approaches such as Q-learning, as the network is updated more rarely and based on more aggregated information, the direct policy search performed by evolutionary algorithms allows in principle for a freer movement in policy space.
Empirically, neuroevolution has been found to reach state-of-the-art performance on reinforcement learning problems which can be solved with small neural networks~\cite{gomez2008accelerated} and to reach close to state-of-the-art performance on games in the ALE benchmark played with visual input~\cite{salimans2017evolution,chrabaszcz2018back}. In general, neuroevolution performs worse in high-dimensional search spaces such as induced by deep neural networks, but there have also been recent results where genetic algorithms have been shown to be competitive with gradient descent for training deep networks for reinforcement learning~\cite{such2017deep}. Neuroevolution has also been found to learn high-performing strategies for a number of other more modern games including racing games and first-person shooters, though using human-constructed features~\cite{risi2017neuroevolution}.

For training the weights of a neural network only, modern variants of evolution strategies can be used. The Covariance Matrix Adaptation Evolution Strategy (CMA-ES)~\cite{hansen2001completely} represents the population implicitly as a distribution of possible search points; it is very effective at training small-size networks in reinforcement learning settings~\cite{igel2003neuroevolution}. Another high-performing development of evolution strategies is the Natural Evolution Strategies (NES) family of algorithms~\cite{wierstra2014natural}. While both CMA and NES suffer from having a number of parameters required for evolution growing superlinearly with the size of the neural network, there are versions that overcome this problem~\cite{schaul2011high,cuccu2012block}.

\subsection{Compressed representation in reinforcement learning}

The high dimensionality of visual input is a problem not only for evolutionary methods, but generally for learning technique. The origin of the success of deep learning can be traced to how deep convolutional networks handle large dimensional inputs; up until a few years ago, reinforcement learning generally relied on low-dimensional features, either by using intrinsically low-dimensional sensors (such as infrared or laser range-finders) or by using hard-coded computer vision techniques to extract low-dimensional state representations from image data. Such hard mappings however do not lend themselves to generalization; in order to create a more general reinforcement learning method, the mapping must be automatically constructed or learned.

Several approaches have been proposed in that sense in reinforcement learning. Some of them rely on neural networks, in particular on various forms of autoencoders~\cite{alvernaz2017autoencoder,ha2018world}. An alternative is to use external compressors such as based on vector quantization~\cite{cuccu2011intrinsically}, where a number of prototype vectors are found and each vector is used as a feature detector--the value of that feature being the similarity between the actual high-dimensional input and the vector, similar to a radial basis function network.

\section{Method}

Our system is divided into four main components: i) the Environment is an Atari game, taking actions and providing observations; ii) the Compressor extracts a low-dimensional code from the observation, while being trained online with the rest of the system; iii) the Controller is our policy approximizer, i.e.\ the neural network; finally iv) the Optimizer is our learning algorithm, improving the performance of the network over time, in our case an Evolution Strategy.
Each component is described in more detail below.

\subsection{Environment}

We test our method on the Arcade Learning Environment (ALE), interfaced through the OpenAI Gym framework~\cite{openaigym}. As discussed above, ALE is built on top of an emulator of the Atari 2600, with all the limitations of that console. In keeping with ALE conventions, the observation consists of a $[210 \times 180 \times 3]$ tensor, representing the RGB pixels of the screen input. The output of the network is interpreted (using one-hot encoding) as one of 18 discrete actions, representing the potential inputs from the joystick. The frame-skipping is fixed at 5 by following each action with 4 \texttt{NOOP} commands.

\subsection{Compressor}
\label{subsec:compressor}

The role of the compressor is to provide a compact representation for each observation coming from the environment, enabling the neural network to entirely focus on decision making.
This is done through unsupervised learning on the very same observations that are obtained by the network interacting with the environment, in an \emph{online learning} fashion.

We address such limitations through a new algorithm based on Vector Quantization (VQ), named Increasing Dictionary VQ, coupled with a new Sparse Coding (SC) method named Direct Residuals SC.
Together they aim at supporting the study of the spaces of observations and features, while offering top performance for online learning.
To the best of our knowledge, the only prior work using unsupervised learning as a pre-processor for neuroevolution is~\cite{cuccu2011intrinsically,alvernaz2017autoencoder}.
The following sections will derive IDVQ+DRSC starting from the vanilla VQ, explaining the design choices which led to these algorithms

\subsubsection{Vanilla vector quantization}

The standard VQ algorithm~\cite{gray1984vector} is a dictionary-based encoding technique with applications in dimensionality reduction and compression.
Representative elements in the space (called singularly \emph{centroids} and collectively called a \emph{dictionary}) act as references for a surrounding volume, in a manner akin to k-means.
The \emph{code} of an element in the space is then a vector where each position corresponds to a centroid in the dictionary.
Its values are traditionally set to zeros, except for the position corresponding to the closest representative centroid in the space.
Variations use a dense code vector, capturing the contribution of multiple centroids for higher precision.
In either case the original can be reconstructed as a vector product between the code and the dictionary.
The difference between the original and its reconstruction is called \emph{reconstruction error}, and quantifies the information lost in the compression/decompression process.
The dictionary is trained by adapting the centroids to minimize reconstruction error over a training set.

Applications to \emph{online} reinforcement learning however present a few limitations.
Additional training data is not only unavailable until late stages, but is also only accessible if obtained by individuals through interaction with the environment.
Take for example an Atari game with different enemies in each level: observing a second-level enemy depends on the ability to solve the first level of the game, requiring in turn the compressor to recognize the first-level enemies.
A successful run should thereby alternate improving the dictionary with improving the candidate solutions: at any stage, the dictionary should provide an encoding supporting the development of sophisticated behavior.

In online learning though, two opposite needs are in play: on one hand, the centroids need to be trained in order to provide a useful and consistent code; on the other hand, late stage training on novel observations requires at least some centroids to be preserved untrained.
Comparing to vanilla VQ, we cannot use random centroids for the code.
As they are uniformly drawn from the space of \emph{all possible images}, their spread is enormously sparse w.r.t.\ the small sub-volume of an Atari game's image.
The similarity of a random centroid to any such image will be about the same: using random centroids as the dictionary consequently produces an \emph{almost constant} code for any image from a same game\footnote{This has also been empirically verified in earlier iterations of this work}.
Image differentiation is relegated to the least significant digits, making it suboptimal as a neural network input.
Directly addressing this problem naturally calls for starting with a smaller dictionary size, and increasing it at later stages as new observations call for it.

%%%%%%%%%%%%%%%%%%%%%%%%%%%%%%%%%%%%%%%%%%%%%%%%%%%%%%%%%%%%%%%%%%%%%
\begin{algorithm}[t!]
  \caption{IDVQ}
  \label{alg:idvq}
  \begin{algorithmic}
    \Inputs{%
      $\trainset$: ~training set, $\orig \in \trainset$  \\
      $\dict$: ~current dictionary\\
      $\delta$: ~~minimal aggregated residual for inclusion\\
    }
    \Initialize{%
      $\dict \gets \varnothing$
        \Comment{dictionary initialized empty}
    }
    \For{$\orig \text{ in } \trainset$}
      \State $\resids \gets \orig$
        \Comment{residual information to encode}
      \State $\code \gets DRSC(\orig, \dict, \epsilon, \Omega)$
        \Comment{$\epsilon$ and $\Omega$ given}
      \State $\recres \gets \code \dict$
      \State $\recerr \gets \resids - \recres$
      \State $\recerrel_i \gets \max(0, \recerrel_i),~\forall \recerrel_i \in \recerr$
        \Comment{remove artifacts}
      \If{$\Sigma|\recerr| > \delta$}
        \State $\dict \append \recerr$
          \Comment{append $\recerr$ to $\dict$}
      \EndIf
    \EndFor\\
    \Return $\dict$
  \end{algorithmic}
\end{algorithm}
%%%%%%%%%%%%%%%%%%%%%%%%%%%%%%%%%%%%%%%%%%%%%%%%%%%%%%%%%%%%%%%%%%%%%

\subsubsection{Increasing Dictionary VQ}

We introduce Increasing Dictionary VQ (IDVQ, Algorithm~\ref{alg:idvq}), a new compressor based on VQ which automatically increases the size of its dictionary over successive training iterations, specifically tailored for online learning.
Rather than having a fixed-size dictionary, IDVQ starts with an \emph{empty} dictionary, thus requiring \emph{no initialization}, and adds new centroids as the learning progresses.

This is done by building new centroids from the positive part of the reconstruction error, which corresponds to the information from the original image (rescaled between 0 and 1) which is not reconstructed by the current encoding (see Algorithm~\ref{alg:idvq}).
Growth in dictionary size is regulated by a threshold $\delta$, indicating the minimal aggregated residual considered to be a meaningful addition.
The training set is built by uniformly sampling the observations obtained by all individuals in a generation.

Centroids added to the dictionary are not further refined.
This is in line with the goal of image differentiation rather than minimizing reconstruction error: each centroid is purposely constructed to represents one particular feature, which was found in an actual observation and was not available in the dictionary before.

Growing the dictionary size however alters the code size, and thus the neural network input size.
This requires careful updates in both the controller and the optimizer, as addressed in Sections~\ref{subsec:controller} and~\ref{subsec:optimizer} respectively.

%%%%%%%%%%%%%%%%%%%%%%%%%%%%%%%%%%%%%%%%%%%%%%%%%%%%%%%%%%%%%%%%%%%%%%%%
\begin{algorithm}[t!]
  \caption{DRSC}
  \label{alg:drsc}
  \begin{algorithmic}
   \Inputs{%
      $\orig$: ~vector to encode (observation)\\
      $\dict$: ~dictionary trained with IDVQ\\
      $\epsilon$: ~~minimal aggregated residual loss\\
      $\maxncodes$: ~maximum nonzero elements in the code
    }
    \Initialize{%
      $\resids \gets X$
        \Comment{residual information to encode}\\
      $\code \gets \vec{0}$
        \Comment{output code}\\
      $\ncodes \gets 0$
        \Comment{non-zero elements in the code}
    }
    \While{$\Sigma|\resids| > \epsilon$ and $\ncodes < \maxncodes$}
      \State $\simils \gets \similfn(\resids, \centr_i), \forall \centr_i \in \dict$
      \State $\msc \gets \maxidx(\simils)$
      \State $\code_\msc \gets 1$
        \Comment{$\code = [\code_1 \dots \code_n]$}
      \State $\ncodes \gets \ncodes + 1$
      \State $\resids \gets \resids - \centr_\msc$
        \Comment{$\dict = [\centr_1 \dots \centr_n]$}
      \State $\resid_i \gets \max (0, \resid_i),~\forall \resid_i \in \resids$
    \EndWhile\\
    \Return $\code$
  \end{algorithmic}
\end{algorithm}
%%%%%%%%%%%%%%%%%%%%%%%%%%%%%%%%%%%%%%%%%%%%%%%%%%%%%%%%%%%%%%%%%%%%%%%%

%%%%%%%%%%%%%%%%%%%%%%%%%%%%%%%%%%%%%%%%%%%%%%%%%%%%%%%%%%%%%%%%%%%%%%%%%%%%%%%%%
\begin{figure*}[!t]
\label{fig:centroids}
\centerline{
  \fbox{\includegraphics[width=0.18\textwidth]{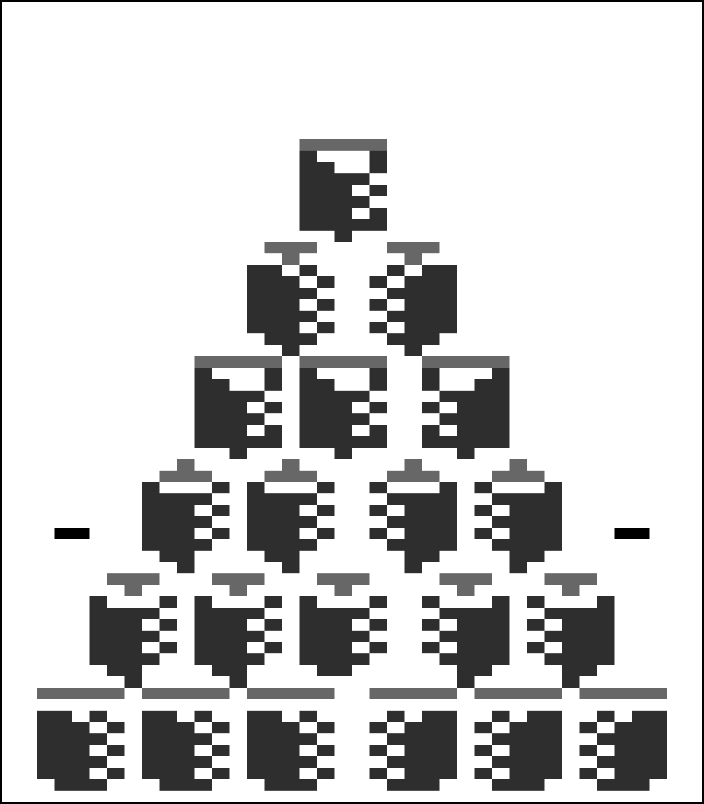}}\hfill
  \fbox{\includegraphics[width=0.18\textwidth]{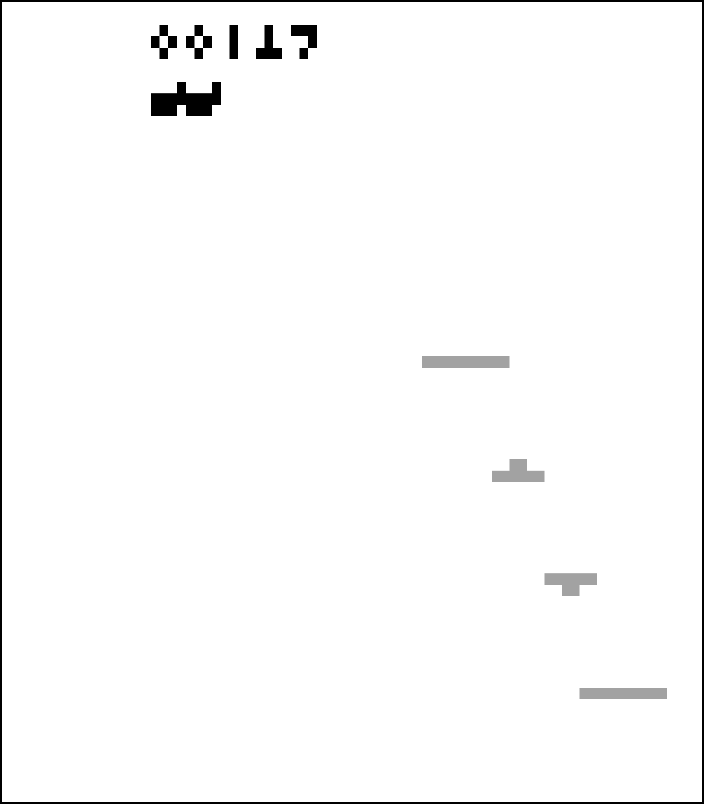}}\hfill
  \fbox{\includegraphics[width=0.18\textwidth]{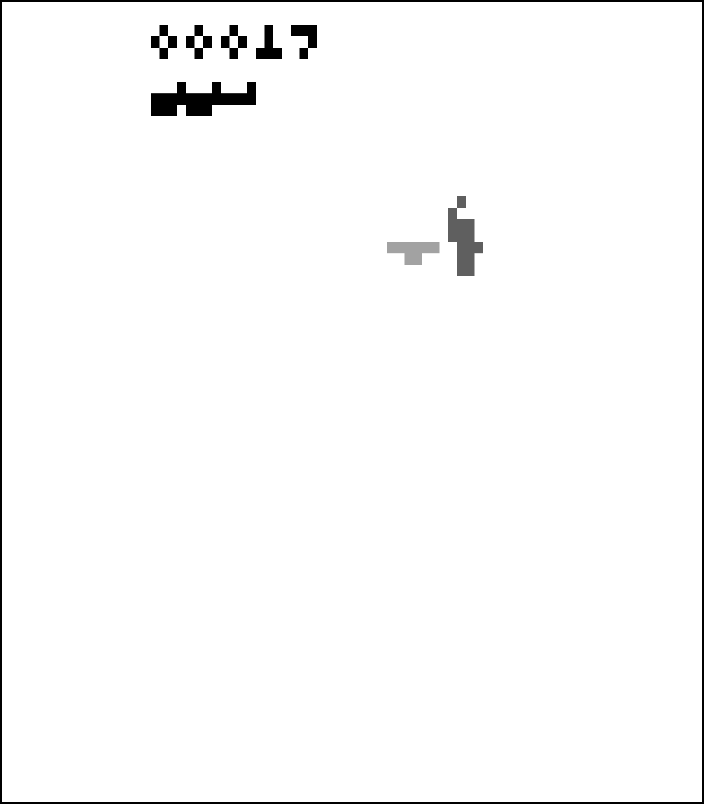}}\hfill
  \fbox{\includegraphics[width=0.18\textwidth]{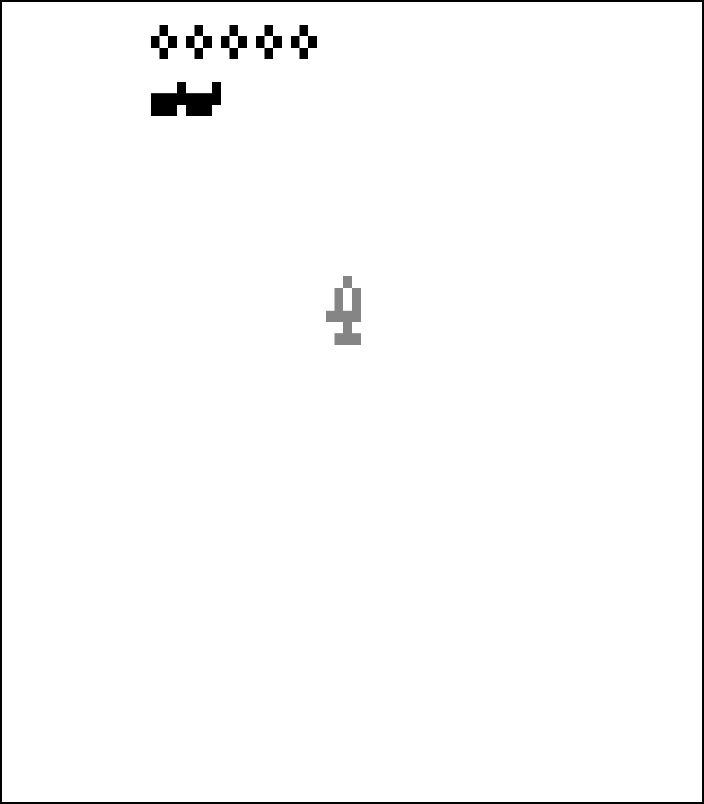}}\hfill
  \fbox{\includegraphics[width=0.18\textwidth]{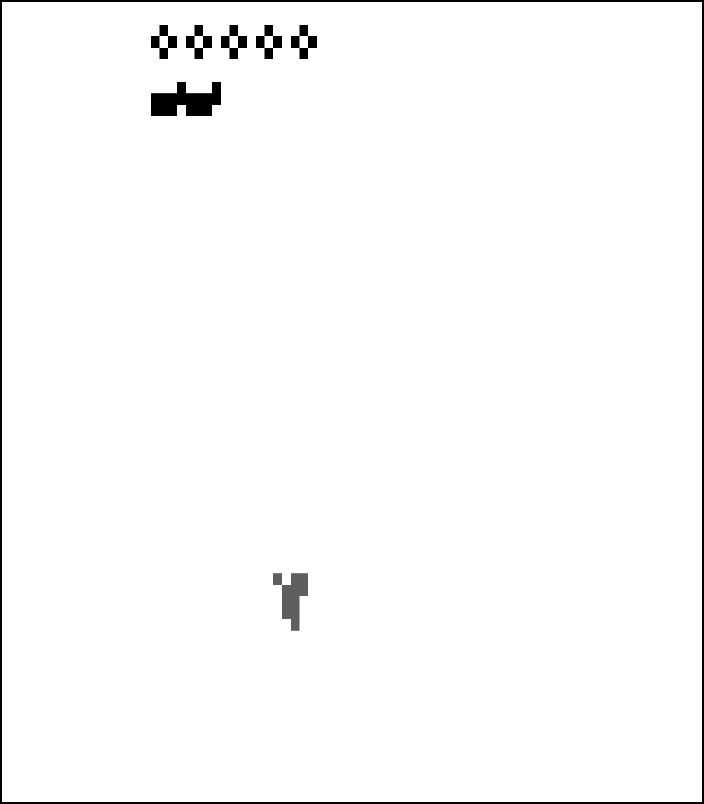}}
}
\caption{
  \textbf{Trained centroids.} A few centroids trained with IDVQ during a run of the game Qbert. Notice how the first captures the initial state of the game (backgroud), while the others build features as subsequent residuals: lit cubes, avatar and enemy. Colors are inverted for printing purposes.
}
\end{figure*}
%%%%%%%%%%%%%%%%%%%%%%%%%%%%%%%%%%%%%%%%%%%%%%%%%%%%%%%%%%%%%%%%%%%%%%%%%%%%%%%%%

\subsubsection{Direct Residuals Sparse Coding}

The performance of algorithms based on dictionary approaches depends more on the choice of encoding than on the dictionary training -- to the point where the best performing algorithms have but a marginal improvement in performance when using sophisticatedly trained centroids versus randomly selected samples~\cite{coates2011importance}.
This highlights the importance of selecting an effective encoding algorithm to best leverage the characteristics of a dictionary trained with IDVQ.
In recent years, several studies have shown algorithms based on \emph{Sparse Coding} to consistently perform best on compression and reconstruction tasks~\cite{mairal2014sparse,zhang2015survey}.
These typically alternate training the centroids and minimizing the $\ell_1$ norm of the code (which approximates $\ell_0$ norm), ultimately yielding a code that is mostly composed of zeros.
In our case though, the dictionary is already trained with IDVQ: we thereby focus on the construction of the sparse code instead.

The classic way to construct a sparse code is through an iterative approach~\cite{mallat1993matching,pati1993orthogonal} where at each step (i) few centroids are selected, (ii) a corresponding code is built and (iii) the code quality is evaluated based on the reconstruction error, with the $\norm_1$ norm of the code as a regularization term.
This process is repeated over different combinations of centroids to incrementally reduce the reconstruction error, at the cost of the algorithm's performance.
Moreover, the reconstruction is computed as a vector product between the code and the dictionary: while conceptually elegant, this dot product produces a linear combination (of the centroids with the code values) where most terms have null coefficients.

In our case though the focus is in differentiating states in order to support the decision maker, rather than perfecting the reconstruction of the original input.
The encoding algorithm will be called on each and every observation coming from the environment, proportionally reducing the computational time available for decision making.
This forces an overhaul of the encoder's objective function from the ground up, prioritizing \emph{distinction over precision}, i.e.\ observation differentiation over reconstruction error.

To this end we introduce Direct Residuals Sparse Coding (DRSC, Algorithm~\ref{alg:drsc}) as a novel sparse coding algorithm specifically tailored to produce highly differentiating encoding in the shortest amount of time.
Its key characteristics are: (i) it utilizes centroids constructed as \emph{residual images} from IDVQ, thus avoiding the centroid-train phase; (ii) it produces binary encodings, reducing the reconstruction process to an unweighted sum over the centroids corresponding to the code's nonzero coefficients; and (iii) it produces the code in a single pass, terminating early after a small number of centroids are selected.
The result is an algorithm with linear performance over dictionary size, which disassembles an observation into its consecutive most similar components as found in the dictionary.

\subsubsection{Step-by-step breakdown}
\label{subsec:compressor:example}

Increasing Dictionary VQ is used to train a dictionary, used by Direct Residuals SC to encode (compress, extract features from) an observation (image).
To understand how these algorithms work together, let us hypothesize a working starting dictionary and see how DRSC produces an encoding.

The initialization includes two steps: the code, as an arrays of zeros with the same size as the dictionary, and the \emph{residual information} still needing encoding, initially the whole original image.
The algorithm then loops to select centroids to add to the encoding, based on how much of the residual information can they encode.
To select the most similar centroid, the algorithm computes the differences between the residual information and each centroid in the dictionary, aggregating each of these differences by summing all values.
The centroid with the smallest aggregated difference is thereby the most similar to the residual information, and is chosen to be included in the encoding.
The corresponding bit in the binary code is flipped to `1', and the residual information is updated by subtracting the new centroid.

The signs of the values in the updated residual information (old residual minus new centroid, the order matters) are now significant:
(i) values equal to zero mean a perfect correspondence between the pixel information in the old residual and the corresponding value in the new centroid;
(ii) positive values correspond to information that was present in the old residual but not covered by the new centroid;
(iii) negative values correspond to information present in the new centroid, but absent (or of smaller magnitude) in the old residual.
This is crucial towards the goal of fully representing the totality of the original information, and to this end the algorithm is free to disregard \emph{reconstruction artifacts} as found in (iii).

Most encoding algorithms make no distinction between not-yet-encoded information and reconstruction artifacts: as they aim at minimizing reconstruction error, they focus on the error's magnitude rather than its origin.
DRSC instead focuses solely on representing all the information initially present in the image, and the artifacts found in the negative values are thereby disregarded by setting them to zero.
The result is a residual image of information present in the original image but not yet captured by the reconstruction.

The algorithm then keeps looping and adding centroids until the (aggregated) residual information is lower than a threshold, corresponding to an arbitrary precision in capturing the information in the original image.
To enforce sparsity in the case that the correct centroids are not available in the dictionary, a secondary stopping criterion for the encoding loop is when too many centroids are added to the code, based on another threshold.
Images with high residual information after encoding are prime candidates for compressor training.

The dictionary is trained with IDVQ by adding new centroids to minimize leftover residual information in the encoding.
The training begins by selecting an image from the training set and encoding it with DRSC, producing the binary code as described above.
A dot product between the code and the dictionary (i.e.\ summing the centroids selected by the code, since it is binary) produces a reconstruction of the original image, similarly to other dictionary-based algorithms.

The difference between the training image and the reconstruction then produces a reconstruction error (-image), where the sign of the values once again correspond to their origin: positive values are leftover information from the image which is not encoded in the reconstruction, while negative values are reconstruction artifacts with no relation to the original image.
This reconstruction error image is then aggregated (with a sum) to estimate the quantity of information missed by the encoding.
If it is above a given threshold, a new centroid should be added to the dictionary to enable DRSC to make a more precise reconstruction.
But in that case the residual itself makes for the perfect centroid, as it exactly captures the information missed by the current encoding, and is then added to the dictionary.

\subsection{Controller}
\label{subsec:controller}

The controller for all experiments is a single-layer fully-connected recurrent neural network (RNN). Each neuron receives the following inputs through weighted connections: the inputs to the network, the output of all neurons from the previous activation (initially zeros), and a constant bias (always set to $1$).
The number of inputs is equal at any given point in time to the size of the code coming from the \emph{compressor}. As the compressor's dictionary grows in size, so does the network's input.
In order to ensure continuity in training (i.e.\ the change needs to be transparent to the training algorithm), it is necessary to define an invariance across this change, where the network with expanded weights is equivalent to the previous one.
This is done by setting the weights of all new connections to zero, making the new network mathematically equivalent to the previous one, as any input on the new connections cancels out.
The same principle can be ported to any neural network application.

The number of neurons in the output layer is kept equal to the dimensionality of the action space for each game, as defined by the ALE simulator.
This is as low as 6 in some games, and 18 at most.
Actions are selected deterministically in correspondence to the maximum activation.
No hidden layer nor extra neurons were used in any of the presented results.
The increase in dimensionality in the input connections' weights corresponds to a growth in the parameter vector of the optimizer, as described below in Section~\ref{subsec:optimizer:dyndim}.

\subsection{Optimizer}
\label{subsec:optimizer}

The optimizer used in the experiments is a variation of Exponential Natural Evolution Strategy(XNES;~\cite{glasmachers2010exponential}) tailored for evolving networks with dynamic varying size.

The next section briefly introduces the base algorithm and its family, followed by details on our modifications.

\subsubsection{Exponential NES}

Natural Evolution Strategies (NES;~\cite{wierstra2008natural,wierstra2014natural}) is a family of evolutionary strategy algorithms that maintain a \emph{distribution} over the parameters space  rather than an explicit population of individuals.
It is distinguishable over similarly distribution-based ES (e.g.\ Covariance Matrix Adaptation Evolution Strategy; CMA-ES~\cite{hansen2001completely}) for its update function based on the \emph{natural} gradient, constructed by rescaling the vanilla gradient based on the Fischer information matrix
$\tilde{\nabla} = \mathbf{F}^{-1} \nabla_\theta J(\theta) $.

The expectation of the fitness function $\fit$ for a given sample $\ind$ with respect to parameters $\theta$ is computed as

\[
  J(\theta) = \mathbb{E}_\theta[\fit(\ind)] = \int \fit(\ind) p(\ind | \theta) d\ind
\]

Where $p(\ind | \theta)$ is a conditional probability distribution function given parameter $\theta$.
This allows writing the updates for the distribution as

\[
  \theta \gets \theta - \eta \tilde{\nabla}_\theta J
    = \theta - \eta \mathbf{F}^{-1} \nabla_\theta J(\theta)
\]

The most representative algorithm of the family is Exponential NES (XNES;~\cite{glasmachers2010exponential}), which maintains a multivariate Gaussian distribution over the parameters space, defined by the parameters $\theta = (\mu, \Sigma)$.
Based on the equation above, with the addition of Monte Carlo estimation, fitness shaping and exponential local coordinates (see~\cite{wierstra2008natural} for the full derivation), these parameters are updated as:

\begin{align*}
  &\mu \gets \mu + \eta_\mu \sum^\lambda_{k=1} u_k \ind_k \\
  &A \gets A \exp(\frac{\eta_A}{2} \sum^\lambda_{k=1} u_k (\ind_k \ind_k^\intercal - \mathcal{I}))
\end{align*}

with $\eta_\mu$ and $\eta_A$ learning rates, $\lambda$ number of estimation samples (the algorithm's correspondent to population size), $u_k$ fitness shaping utilities, and $A$ upper triangular matrix from the Choleski decomposition of $\Sigma$, $\Sigma = A^\intercal A$.

The update equation for $\Sigma$ bounds the performance to $\order(\nparams^3)$ with $\nparams$ number of parameters.
At the time of its inception, this limited XNES to applications of few hundred dimensions.
Leveraging modern hardware and libraries though, our current implementation easily runs on several thousands of parameters in minutes%
\footnote{For a NES algorithm suitable for evolving deep neural networks see Block Diagonal NES~\cite{cuccu2012block}, which scales linearly on the number of neurons / layers.}%
.
% Perhaps more importantly, its parametrization makes it a prime candidate for a GPU-based implementation, as long as $\theta$ can be maintained on GPU memory, with only the individuals being fetched at each generation.

%%%%%%%%%%%%%%%%%%%%%%%%%%%%%%%%%%%%%%%%%%%%%%%%%%%%%%%%%%%%%%%%%%%%%%%%%%%%%%%%%%
\begin{table*}[t!]
\caption{%
\textbf{Game scores.} Scores on a sample of Atari games (sorted alphabetically), compared to results from HyperNeat~\citep{hausknecht2014neuroevolution} and OpenAI ES~\citep{salimans2017evolution}.
Results from GA (1B)~\citep{such2017deep} and NSRA-ES~\citep{conti2018improving} are also provided (though the intersection between games sets is minimal) to include work aimed at \emph{expanding} the network size, rather than shrinking it.
All methods were trained from scratch on raw pixel input (NSRA-ES uses a compact state representation read from the simulated Atari RAM to compute novelty).
Column `\emph{\# of neurons}' indicates how many neurons were used in our work in a single layer (output) for each game. The number of neurons corresponds to the number of available actions in each game, i.e.\ no neurons are added for performance purpose.
}

\label{tab:scores}
\center

\begin{tabular}{m{25mm}llllll}

\toprule
Game           & HyperNeat & OpenAI ES & GA (1B) & NSRA-ES  & \textbf{IDVQ+DRSC+XNES}  & \# of neurons  \\
\midrule
DemonAttack    & 3590      & 1166.5    & -       & -       & 325      & 6      \\
FishingDerby   & -49       & -49       & -       & -       & -10      & 18     \\
Frostbite      & 2260      & 370       & 4536    & 3785    & 300      & 18     \\
Kangaroo       & 800       & 11200     & 3790    & -       & 1200     & 18     \\
NameThisGame   & 6742      & 4503      & -       & -       & 920      & 6      \\
Phoenix        & 1762      & 4041      & -       & -       & 4600     & 8      \\
Qbert          & 695       & 147.5     & -       & 1350    & 1250     & 6      \\
Seaquest       & 716       & 1390      & 798     & 960     & 320      & 18     \\
SpaceInvaders  & 1251      & 678.5     & -       & -       & 830      & 6      \\
TimePilot      & 7340      & 4970      & -       & -       & 4600     & 10     \\
\bottomrule

\end{tabular}
\end{table*}
%%%%%%%%%%%%%%%%%%%%%%%%%%%%%%%%%%%%%%%%%%%%%%%%%%%%%%%%%%%%%%%%%%%%%%%%%%%%%%%%%%
\begin{table*}[t!]
\caption{%
  \textbf{Results.}
  Our proposed approach achieves comparable scores (sometimes better) using up to \emph{two orders of magnitude} less neurons, and no hidden layers.
  The proposed feature extraction algorithm IDVQ+DRSC is simple enough (using basic, linear operations) to be arguably unable to contribute to the decision making process in a sensible manner (see Section~\ref{subsec:compressor:example}). This implies that the tiny network trained on decision making alone is of sufficient complexity to learn a successful policy, potentially prompting for reconsidering the actual complexity of this standard benchmark. The following numbers refer to networks for games with the largest action set (18). See Table~\ref{tab:scores} for the actual number of neurons used in the output layer for each game.
}
\label{tab:nneurs}
\center
\begin{tabular}{m{25mm}lllll}
  \toprule
  & HyperNeat & OpenAI ES & GA (1B) & NSRA-ES & \textbf{IDVQ+DRSC+XNES} \\
  \midrule
  \# neurons       & {\textasciitilde{}}3034 & {\textasciitilde{}}650 & {\textasciitilde{}}650  & {\textasciitilde{}}650  & \textbf{{\textasciitilde{}}18} \\
  \# hidden layers & 2 & 3 & 3 & 3 & \textbf{0} \\
  \# connections   & {\textasciitilde{}}906k & {\textasciitilde{}}436k & {\textasciitilde{}}436k & {\textasciitilde{}}436k & \textbf{{\textasciitilde{}}}\textbf{3k}\\
  \bottomrule
  \end{tabular}
\end{table*}
%%%%%%%%%%%%%%%%%%%%%%%%%%%%%%%%%%%%%%%%%%%%%%%%%%%%%%%%%%%%%%%%%%%%%%%%%%%%%%%%%%

\subsubsection{Dynamically varying the dimensionality}
\label{subsec:optimizer:dyndim}

This paper introduces a novel twist to the algorithm as the dimensionality of the distribution (and thus its parameters) varies during the run.
Since the parameters are interpreted as network weights in \emph{direct encoding} neuroevolution, changes in the network structure need to be reflected by the optimizer in order for future samples to include the new weights.
Particularly, the multivariate Gaussian acquires new dimensions: $\theta$ should be updated keeping into account the order in which the coefficients of the distribution samples are inserted in the network topology.

In Section~\ref{subsec:controller} we explain how the network update is carried through by initializing the new weights to zeros.
In order to respect the network's invariance, the expected value of the distribution ($\mu$) for the new dimension should be zero.
As for $\Sigma$, we need values for the new rows and columns in correspondence to the new dimensions.
We know that (i) the new weights did not vary so far in relation to the others (as they were equivalent to being fixed to zero until now), and that (ii) everything learned by the algorithm until now was based on the samples having always zeros in these positions.
So $\Sigma$ must have for all new dimensions (i) zeros covariance and (ii) arbitrarily small variance (diagonal), only in order to bootstrap the search along these new dimensions.

Take for example a one-neuron feed-forward network with 2 inputs plus bias, totaling 3 weights.
Let us select a function mapping the optimizer's parameters to the weights in the network structure (i.e.\ the \emph{genotype to phenotype} function), as to first fill the values of all input connections, then all bias connections.
Extending the input size to 4 requires the optimizer to consider two more weights before filling in the bias:

\begin{align*}
\mu &=
  &&\begin{bmatrix}
    \mu_1 & \mu_2 & \mu_b
  \end{bmatrix}
  && \rightarrow &&
  \begin{bmatrix}
    \mu_1 & \mu_2 & 0 & 0 & \mu_b
  \end{bmatrix}\\
\Sigma &=
  &&\begin{bmatrix}
      \sigma^2_1 & \cov_{12} & \cov_{1b} \\
      \cov_{21} & \sigma^2_2 & \cov_{2b} \\
      \cov_{b1} & \cov_{b2} & \sigma^2_b
  \end{bmatrix}
  && \rightarrow &&
  \begin{bmatrix}
      \sigma^2_1 & \cov_{12} & 0 & 0 & \cov_{1b} \\
      \cov_{21} & \sigma^2_2 & 0 & 0 & \cov_{2b} \\
      0 & 0 & \epsilon & 0 & 0 \\
      0 & 0 & 0 & \epsilon & 0 \\
      \cov_{b1} & \cov_{b2} & 0 & 0 & \sigma^2_b
  \end{bmatrix}
\end{align*}

% \vspace{-1ex}

with $\cov_{ij}$ being the covariance between parameters $i$ and $j$, $\sigma^2_k$ the variance on parameter $k$, and $\epsilon$ being arbitrarily small ($0.0001$ here).
The complexity of this step of course increases considerably with more sophisticated mappings, for example when accounting for recurrent connections and multiple neurons, but the basic idea stays the same.
The evolution can pick up from this point on as if simply resuming, and learn how the new parameters influence the fitness.

\section{Experimental setup}

The experimental setup further highlights the performance gain achieved, and is thus crucial to properly understand the results presented in the next section:

\begin{compactenum}[$\bullet$]
  % \item Code is written from scratch in Ruby and available open source on GitHub\footnote{Experiments: \url{github.com/giuse/DNE}; algorithms: \url{github.com/giuse/machine_learning_workbench}.}.
  \item All experiments were run on a single machine, using a 32-core Intel(R) Xeon(R) E5-2620 at 2.10GHz, with only 3GB of ram per core (including the Atari simulator and Python wrapper).
  \item The maximum run length on all games is capped to $200$ interactions, meaning the agents are alloted a mere $1'000$ frames, given our constant frameskip of 5. This was done to limit the run time, but in most games longer runs correspond to higher scores.
  \item Population size and learning rates are dynamically adjusted based on the number of parameters, based on the XNES minimal population size and default learning rate~\cite{glasmachers2010exponential}. We scale the population size by $1.5$ and the learning rate by $0.5$. In all runs on all games, the population size is between $18$ and $42$, again very limited in order to optimize run time on the available hardware.
  \item The dictionary growth is roughly controlled by $\delta$ (see Algorithm~\ref{alg:idvq}), but depends on the graphics of each game. The average dictionary size by the end of the run is around 30-50 centroids, but games with many small moving parts tend to grow over 100. In such games there seems to be direct correlation between higher dictionary size and performance, but our reference machine performed poorly over 150 centroids. We found numbers close to $\delta = 0.005$ to be robust in our setup across all games.
  \item Graphics resolution is reduced from $[210 \times 180 \times 3]$ to $[70 \times 80]$, averaging the color channels to obtain a grayscale image. This also contributes to lower run times.
  \item Every individual is evaluated 5 times to reduce fitness variance.
  \item Experiments are allotted a mere 100 generations, which averages to 2 to 3 hours of run time on our reference machine.
\end{compactenum}

These computational restrictions are \textbf{extremely tight} compared to what is typically used in studies utilizing the ALE framework.
Limited experimentation indicates that relaxing any of them, i.e.\ by accessing the kind of hardware usually dedicated to modern deep learning, consistently improves the results on the presented games.
The full implementation is available on GitHub under MIT license\footnote{\url{https://github.com/giuse/DNE/tree/six_neurons}}.

\section{Results}

The goal of this work is not to propose a new generic feature extractor for Atari games, nor a novel approach to beat the best scores from the literature.
Our declared goal is to show that dividing feature extraction from decision making enables tackling hard problems with minimal resources and simplistic methods, and that the deep networks typically dedicated to this task can be substituted for simple encoders and tiny networks while maintaining comparable performance.
Table~\ref{tab:nneurs} emphasizes our findings in this regard.

Under these assumptions, Table~\ref{tab:scores} presents comparative results over a set of 10 Atari games from the hundreds available on the ALE simulator.
This selection is the result of the following filtering steps: (i) games available through the OpenAI Gym; (ii) games with the same observation resolution of $[210, 160]$ (simply for implementation purposes); (iii) games not involving 3D perspective (to simplify the feature extractor).
The resulting list was further narrowed down due to hardware and runtime limitations.
A broader selection of games would support a broader applicability of our particular, specialized setup; our work on the other hand aims at highlighting that our simple setup is indeed able to play Atari games with competitive results.

Results on each game differ depending on the hyperparameter setup.
To offer a more direct comparison, we opted for using the same settings as described above for all games, rather than specializing the parameters for each game.
Some games performed well with these parameters (e.g.\ Phoenix); others feature many small moving parts in the observations, which would require a larger number of centroids for a proper encoding (e.g.\ Name This Game, Kangaroo); still others have complex dynamics, difficult to learn with such tiny networks (e.g.\ Demon Attack, Seaquest).

The resulting scores are compared with recent papers that offer a broad set of results across Atari games on comparable settings, namely \cite{salimans2017evolution,such2017deep,conti2018improving,hausknecht2014neuroevolution}.
Our list of games and correspondent results are available in Table~\ref{tab:scores}.
Notably, our setup achieves high scores on \emph{Qbert}, arguably one of the harder games for its requirement of strategic planning.

The real results of the paper however are highlighted in Table~\ref{tab:nneurs}, which compares the number of neurons, hidden layers and total connections utilized by each approach.
Our setup uses up to two order of magnitude less neurons, two orders of magnitude less connections, and is the only one using only one layer (no hidden).

\section{Conclusions}

We presented a method to address complex learning tasks such as learning to play Atari games by decoupling policy learning from feature construction, learning them independently but simultaneously to further specializes each role.
Features are extracted from raw pixel observations coming from the game using a novel and efficient sparse coding algorithm named Direct Residual Sparse Coding.
The resulting compact code is based on a dictionary trained online with yet another new algorithm called Increasing Dictionary Vector Quantization, which uses the observations obtained by the networks' interactions with the environment as the policy search progresses.
Finally, tiny neural networks are evolved to decide actions based on the encoded observations, to achieving results comparable with the deep neural networks typically used for these problems while being \emph{two orders of magnitude} smaller.

Our work shows how a relatively simple and efficient feature extraction method, which counter-intuitively does not use reconstruction error for training, can effectively extract meaningful features from a range of different games.
The implication is that feature extraction on some Atari games is not as complex as often considered.
On top of that, the neural network trained for policy approximation is also very small in size, showing that the decision making itself can be done by relatively simple functions.

We empirically evaluated our method on a set of well-known Atari games using the ALE benchmark.
Tight performance restrictions are posed on these evaluations, which can run on common personal computing hardware as opposed to the large server farms often used for deep reinforcement learning research.
The source code is open sourced for further reproducibility.
The game scores are in line with the state of the art in neuroevolution, while using but a minimal fraction of the computational resources usually devoted to this task.
One goal of this paper is to clear the way for new approaches to learning, and to call into question a certain orthodoxy in deep reinforcement learning, namely that image processing and policy should be learned together \emph{(end-to-end)}.

As future work, we plan to identifying the actual complexity required to achieve top scores on a (broader) set of games.
This requires first applying a feature extraction method with state-of-the-art performance, such as based on \emph{autoencoders}.
Our findings though support the design of novel variations focused on state differentiation rather than reconstruction error minimization.
As for the decision maker, the natural next step is to train deep networks entirely dedicated to policy learning, capable in principle of scaling to problems of unprecedented complexity.
Training large, complex networks with neuroevolution requires further investigation in scaling sophisticated evolutionary algorithms to higher dimensions.
An alternative research direction considers the application of deep reinforcement learning methods on top of the external feature extractor.
Finally a straightforward direction to improve scores is simply to release the constraints on available performance: longer runs, optimized code and parallelization should still find room for improvement even using our current, minimal setup.

\section*{Acknowledgments}

We kindly thank Somayeh Danafar for her contribution to the discussions which eventually led to the design of the IDVQ and DRSC algorithms.

\normalsize
\bibliography{bibliography}
\balance % BALANCE REFERENCE COLUMNS!!

\end{document}